\documentclass[11pt]{amsart}
\usepackage[a4paper, left=3cm,top=3cm,bottom=3cm,right=3cm]{geometry}
\usepackage{natbib}
\usepackage{graphicx,amssymb, amsmath,amsthm}
\usepackage{amsaddr}
\newcommand{\be}{\b \eta}
\newcommand{\bt}{\b \theta}
\newcommand{\bl}{\b \lambda}
\newcommand{\T}{^{\scriptstyle\mathrm{T}}}
\renewcommand{\b}[1]{\boldsymbol{#1}}
\newcommand{\diag}{\mathrm{diag}}
\newcommand{\ci}{\mbox{\protect $\: \perp \hspace{-2.3ex}\perp$ }}

\newcommand{\pcal}{\mathcal{P}}
\newcommand{\dcal}{\mathcal{D}}

\newcommand{\mcal}{\mathcal{M}}
\newcommand{\lcal}{\mathcal{L}}

\newcommand{\ccal}{\mathcal{C}}
\newcommand{\qcal}{\mathcal{Q}}

\newcommand{\bitem}{\begin{itemize}}
\newcommand{\eitem}{\end{itemize}}
\newcommand{\benum}{\begin{enumerate}}
\newcommand{\eenum}{\end{enumerate}}

\theoremstyle{plain}

\newtheorem{prop}{Proposition}
\newtheorem{defn}{Definition}[section]
\newtheorem{lemma}{Lemma}

\newtheoremstyle{example}
  {10pt}
  {10pt}
  {}
  {}
  {\bfseries}
  {.}
  {.5em}
  {}

\theoremstyle{example}
\newtheorem{example}{Example}
\newcommand{\bex}{\begin{example}}
\newcommand{\eex}{\end{example}}

\title[Bi-directed graph models for categorical data]{Parameterizations and fitting of bi-directed graph models to categorical data}
\author{Monia Lupparelli}
\address{Dipartimento di Economia Politica e Metodi Quantitativi,\\via S. Felice, 7, 27100, Pavia, Italy}
\email{mlupparelli@eco.unipv.it}
\author{Giovanni M. Marchetti}
\email{giovanni.marchetti@ds.unifi.it}
\address{Dipartimento di Statistica ``G. Parenti'',\\ viale Morgagni, 59, 50134, Florence, Italy}
\author{Wicher P. Bergsma}
\address{London School of Economics and Political Science,\\ Houghton Street, WC2A 2AE London, UK}
\email{W.P.Bergsma@lse.ac.uk}
\date{3 January 2008}
\begin{document}
\begin{abstract}
We discuss  two parameterizations of models for marginal independencies for  discrete distributions  which are representable by
bi-directed graph models, under the global Markov property. Such models are useful data analytic tools especially if used in combination with
other  graphical models. The first parameterization, in the saturated case,
 is  also known as the multivariate logistic transformation, the second is a variant that allows,
 in some  (but not all) cases, variation independent parameters. An
algorithm for maximum likelihood fitting is proposed, based  on an extension of
the Aitchison and Silvey  method.
\end{abstract}


\maketitle
\newpage
\section{Introduction}
This paper deals with the parametrization and fitting of a class
of marginal independence models for multivariate discrete
distributions. These models are associated to a class of graphs where the
missing edges represent marginal independence. The graphs used
have special edges  to  distinguish them from  undirected graphs
used to encode conditional independencies.  \citet{cw:1993} use
dashed edges and call the graphs covariance graphs by stressing
the equivalence between a marginal pairwise independence and a
zero covariance in a Gaussian distribution. \citet{ric-spi:2002}
use instead bi-directed edges following the tradition of path
analysts. The interpretation of the graphs in terms of independencies
is based on the   pairwise and global Markov properties
discussed originally by \citet{kau:1996} for covariance graphs and later developed by
\citet{ric:2003}. These are recalled in Section 2.

\begin{figure}[b]
\centering
\begin{tabular}{ccc}
  \includegraphics[height=2.4cm]{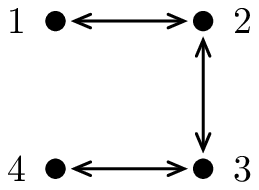} &
 \includegraphics[height=2.4cm]{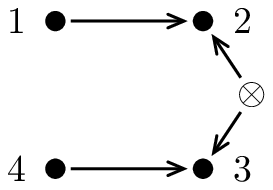} &
 \includegraphics[height=2.4cm]{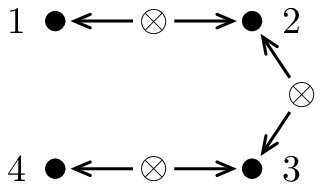} \\
(a) & (b) & (c) \\
\end{tabular}
\caption{\it (a) A bi-directed graph, called 4-chain,  implying the independencies: $4 \ci 12$ and $1
\ci 34$. Directed acyclic graphs  inducing the same independencies after marginalization
over the latent variables (with nodes  $\otimes$): (b)  with one latent variable; (c) with 3 latent variables. }
 \label{fig.introd}
\end{figure}

Models of marginal independence can be useful in several contexts.
For instance, \citet{cw:1993} present an example on diabetic patients
concerning  four continuous variables: $X_1$, the duration of the illness, $X_2$,
the quantity of a particular metabolic parameter, $X_3$, a score for the knowledge about the
illness,  and $X_4$, a questionnaire score measuring a patients'
attitude called external fatalism. The structure of the correlation matrix
suggests for this  data set the marginal independencies $X_4 \ci \{X_1,X_2\}$
and $X_1 \ci \{X_3,X_4\}$. This marginal
independence model can be represented by the bi-directed graph in
Figure~\ref{fig.introd}(a), called a  4-chain. The suggested interpretation is that
the duration of illness $X_1$ and the external fatalism $X_4$ are
independent explanatory variables of the responses $X_2,X_3$ in two seemingly
unrelated regressions. For further discussion on the interpretation of covariance chains
see \citep{wer-cox-mar:2006}.
Bi-directed graph models are sometimes useful to represent marginal
independence structures induced after marginalizing over latent
variables. The independence structure of the diabetes data, for example,
might be represented by assuming an underlying  generating process described by  a
 directed acyclic graph, shown in Figure~\ref{fig.introd}(b),
with one latent variable  pointing both to $X_2$ and $X_3$. After
marginalizing over the latent variable the
induced  independencies are  exactly those encoded in the bi-directed graph of Figure~\ref{fig.introd}(a).
\begin{table}[b]
\centering
\caption{\it Data by \citet{coppen:1966} on symptoms of psychiatric patients. The variables are
 $X_1:$ stability (1=extroverted, 2=introverted), $X_2:$  validity
(1=psychasthenic, 2=energetic),
$X_3:$  depression (yes, no), $X_4:$ solidity (1=hysteric, 2=rigid).}
\label{tab.coppen}
\begin{small}
\begin{tabular}{lllrrrr} \hline
        &    & $X_4$ &         1&             &         2&              \\
$X_1$       &  $X_3$ & $X_2$ &         1&            2&         1&            2 \\ \hline
1       &  y &   &       15 &          30 &        9 &          32  \\
        &  n &   &       25 &          22 &       46 &          27  \\
2       &  y &   &       23 &          22 &       14 &          16  \\
        &  n &   &       14 &           8 &       47 &          12  \\ \hline
\end{tabular}
\end{small}
\end{table}
As another example with four binary variables, consider the data
by \citet{coppen:1966} shown in Table~\ref{tab.coppen}, concerning
symptoms of 362 psychiatric patients. The symptoms are: $X_1:$
stability,  $X_2:$ validity, $X_3:$ acute depression and   $X_4:$
solidity. The chi-squared tests of the hypotheses of marginal
independence   $X_4 \ci \{X_1, X_2\}$ and $X_1 \ci \{X_3, X_4\}$,
with p-values, respectively, $0.32$ and $0.14$, are separately not
significant and the independence model defined by the two
statements jointly gives a satisfactory fit with a deviance of
$8.61$ on $5$ degrees of freedom. Thus the same bi-directed graph
model defined by the 4-chain of Figure~\ref{fig.introd}(a) is
adequate. In Section~\ref{sec.app} we discuss the details of this
application. In this example, if all  symptoms are treated on the
same footing, it is less plausible that a single latent variable
will explain the independence structure and more (at least three)
latent variables are required to suggest a generating process, as
shown in the graph of Figure~\ref{fig.introd}(c).

Developing a parameterization for Gaussian bi-directed  graph
models is straightforward since the pairwise and the global Markov
property are equivalent and they can be simply fulfilled by
constraining to zero a subset of covariances.  Accomplishing the same task
in the discrete case is much more difficult due to
the high number of  parameters and to the
non-equivalence of the two Markov properties.
Recently, \citet{drt-ric:2007} studied the parametrization of
bi-directed graph models for discrete binary
distributions, based on M\"{o}ebius  parameters, by proposing
a version of their iterative conditional fitting algorithm
for maximum likelihood estimation.

In this paper we propose different parameterizations, suitable for
general categorical variables, based on the class of marginal
log-linear models of \citet{ber-rud:2002}. One special case of
this class, especially useful in the context of bi-directed graph
models, is the multivariate logistic parameterization of
\cite{glo-mcc:1995}; see also \cite{kau:1997}. We discuss  a
further marginal log-linear parametrization that can, in special
cases, be shown to imply variation independent parameters. We show
that the marginal log-linear parameterizations suggest a class of
reduced models defined by constraining  certain higher-order
log-linear parameters to zero. Then we discuss maximum likelihood
estimation of the models and we propose a general algorithm based
on previous works by \cite{ait-sil:1958}, \cite{lang:1996},
\cite{ber:1997}.

The remainder of this paper is organized as follows. Section
\ref{sec.bid} reviews discrete bi-directed graphs and their
Markov properties. In Section \ref{sec.mar}  we give the essential results concerning the theory of marginal log-linear models.
Two parameterizations of bi-directed graph models are  given then in Section~\ref{sec.par} illustrating their properties
with special emphasis on  variation independence and the interpretation
  of the parameters.
In Section \ref{sec.fit} we propose an algorithm for   maximum likelihood  fitting and then, in
  Section~\ref{sec.app} we provide some examples. Finally, in Section~\ref{sec.dis}  we give a short discussion, with
a comparison with the approach by \citet{drt-ric:2007}.

\section{Discrete bi-directed graph models}\label{sec.bid}
Bi-directed graphs are essentially undirected graphs with edges
represented by bi-directed arrows instead of full lines. We review
in this section the main concepts of graph theory required to
understand the models. A bi-directed graph $G=(V, E)$ is a pair $G
= (V, E)$, where $V = \{1, \dots, d\}$ is a set of nodes, and $E$
is a set of edges  defined by  two-element subsets of $V$. Two
nodes $u,v$  are \emph{adjacent} or neighbours if $uv$ is an edge
of $G$ and  in this case the edge is drawn as bi-directed, $u
\longleftrightarrow v$. Two edges are adjacent if they have an end
node in common. A \emph{path} from a node $u$ to a node $v$ is a
sequence of adjacent edges connecting $u$ and $v$ for which the
corresponding sequence of nodes contains no repetitions. The nodes
$u$ and $v$ are called  the \emph{endpoints} of the path and all
the other nodes are called  the \emph{inner nodes}.

A graph $G$ is \emph{complete} if all its nodes are pairwise
adjacent. A non-empty graph $G$ is called \emph{connected} if any
two of its nodes are linked by a path in $G$, otherwise it is
called \emph{disconnected}. If $A$ is a subset of the node set $V$
of $G$, the graph $G_A$ with nodes $A$ and containing all the
edges of $G$ with endpoints in $A$ is called an \emph{induced
subgraph}. If a subgraph $G_A$ is connected (resp. disconnected,
complete) we call also $A$ connected (resp. disconnected,
complete), in $G$. The set of all disconnected sets of the graph
$G$ will be denoted by $\dcal$, and the set of all the connected
sets of $G$ will be denoted by $\ccal$. In a graph $G$ a
\emph{connected component} or simply a component is a maximal
connected subgraph. If a subset $D$ of nodes is disconnected then
it  can be uniquely decomposed into more connected components
$C_1, \dots, C_r$, say, such that $D = C_1\cup \cdots\cup C_r$.

The usual notion of separation in undirected graphs  can be used
also for bi-directed graphs. Thus, given three disjoint subsets of
nodes $A$, $B$ and $C$,  $A$ and $B$ are said to be
\emph{separated by} $C$ if for any $u$ in $A$ and any $v$ in $B$
all paths from $u$ to $v$ have at least one inner node in $C$. The
cardinality of a set $V$  will be denoted by $|V|$. The set of all
the subsets of $V$, the power set, will be denoted by $\pcal(V)$.
We use also the notation $\pcal_0(V)$ for the set of all nonempty
subsets of $V$.

Let $X =(X_v, v \in V)$ be  a discrete random vector with each component
$X_v$ taking on values in the finite set  $\mathcal{I}_v = \{1, \dots, b_v\}$.
The Cartesian product $\mathcal{I}_V = \times_{v\in V} \mathcal{I}_v$,
is a contingency table, with generic element  $\b i = (i_v, v \in V)$, called a cell of the table,
and with total number of cells  $t=|\mathcal{I}_V|$.
We assume that $X$ has a joint  probability function $p(\b i)$, $\b i \in \mathcal{I}_V$
giving the probability that an individual falls in cell $\b i$.
Given  a subset $M\subseteq V$ of the variables, the  marginal contingency table is
$\mathcal{I}_M = \times_{v\in M} \mathcal{I}_v$ with generic cell $\b i_M$ and the marginal probability function
of the random vector $X_M = (X_v, v \in M)$ is
$p_M(\b i_M) = \sum_{\b j  \in \mathcal{I}_V|  \b j_M = \b i_M} p(\b j)$.

A bi-directed graph $G = (V, E)$ induces an independence model for
the discrete random vector $X=(X_v, v \in V)$ by defining a Markov
property, i.e. a rule for reading off the graph the
independence relations. In the following we shall use the shorthand notation  $A \ci B | C$
to indicate the conditional independence $X_A \ci X_B | X_C$, where  $A$, $B$ and $C$  are three disjoint
subsets of $V$. Similarly $A \ci B$ and $A \ci B \ci C$ will denote
the marginal and the complete independence, respectively,  of sub-vectors of $X$.
There are two  Markov properties describing the
independence model associated with  a bi-directed
graph, which we consider in this paper:
(a) the global Markov property of \citet{kau:1996} and (b) the connected set Markov property  by \citet{ric:2003}.

The distribution of the random vector $X$ satisfies the \emph{global Markov property} for the bi-directed graph $G$ if
for any triple of disjoint sets $A$, $B$ and $C$,
$$
A  \ci B \mid V \setminus (A \cup B \cup C) \text{ whenever } A  \text{  is separated from } B \text{ by } C \text{ in } G.
$$
Instead, the distribution of $X$ is said to satisfy the \emph{connected set Markov property} if
\begin{multline} \label{eq.conn}
C_1 \ci   \cdots \ci  C_r  \text{ whenever } C_1, \dots, C_r  \text{ are  the connected components}\\
\text{ of every disconnected set } D \in \dcal.
\end{multline}
\citet{ric:2003}  proves that the two properties are equivalent;
see also \citet{drt-ric:2007}. Following these authors
we define a discrete bi-directed graph model as follows.
\begin{defn}\label{def:dbgm}
A \emph{discrete bi-directed graph model} associated with a bi-directed graph $G=(V, E)$ is a
family of discrete joint probability distributions $p$ for the
discrete random vector $X = (X_v, v \in V)$,  that satisfies the property \eqref{eq.conn} for $G$, i.e.
such that, for every disconnected set $D$ in the graph,
$$
p_D(\b i_D) = p_{C_1}(\b i_{C_1}) \times \cdots \times
p_{C_r}(\b i_{C_r}),
$$
where   $C_1, \dots, C_r$ are the connected components of $D$.
\end{defn}
If the global Markov property holds then for any pair  of not
adjacent nodes, the associated random variables are marginally
independent. This implication is called the \emph{pairwise Markov
property} and it is for discrete variables a necessary but not
sufficient condition  for the global Markov property. This is in
sharp contrast with the family of Gaussian distributions where the
two properties are equivalent.

\begin{figure}[t]
 \centering
\begin{tabular}{cc}
\includegraphics[height = 2.8cm]{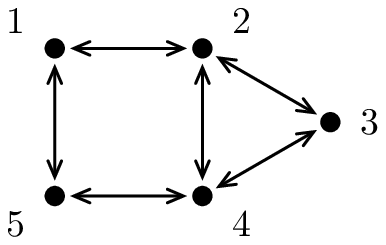} &
\includegraphics[height = 2.8cm]{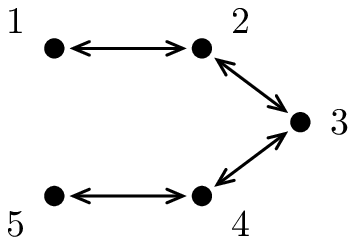} \\
(a) & (b) \\
\end{tabular}
\caption{\it Two bi-directed graphs. The independencies
implied by the connected set Markov property (or, equivalently,
the global Markov property)  are: {\rm (a)} $1 \ci 34$, $3\ci 15$
and $5 \ci 23$;  {\rm (b)} $1 \ci 3 \ci 5$,  $1 \ci 345$, $12 \ci
45$ and $123 \ci 5$.} \label{fig.sec2}
\end{figure}

\begin{example}\label{ex.glob}
Here and henceforth we shall use the short forms $34$ and $12$ to denote
the sets $\{3,4\}$ and $\{1,2\}$, and so on.
The graph of
Figure~\ref{fig.introd}(a) is a chain in 4 nodes with disconnected sets
$$
\dcal = \{13, 14, 24, 134, 124\}.
$$
Thus, $D = 13$ has  the components $C_1 = 1$ and $C_2 = 3$, while
$D = 134$ can be decomposed into $C_1 = 1$ and $C_2 = 34$.
The pairwise Markov property implies $1
\ci 3$,  $1 \ci 4$ and    $2 \ci 4$, while  the connected set  Markov
property implies further that $1 \ci 34$ and $4 \ci 12$. The
global Markov property implies the equivalent set of
independence statements $1 \ci 4$, $ 2\ci 4| 1$ and
$1 \ci 3|4$.
\end{example}
Note that  the complete list of all marginal independencies
implied by a bi-directed graph model is derived from
the class $\dcal$ of all disconnected sets of the graph.
\begin{example}
The graph of Figure~\ref{fig.sec2}(a) has 7 disconnected sets and thus the associated discrete
bi-directed graph model fulfills the independencies
$$
1 \ci 3, \; 1 \ci 4, \; 2 \ci 5, \; 3 \ci 5, \; 1 \ci 34, \;
5 \ci 23, \; 3 \ci 15
$$
that reduce to   $1 \ci 34$, $3 \ci 15$ and $5 \ci 23$, after eliminating redundancies.
The discrete model associated with the graph of Figure~\ref{fig.sec2}(b) with  16 disconnected subsets
satisfies 16  marginal independencies that
can be reduced to the four statements
$$
1 \ci 3 \ci 5, \quad 1 \ci 345, \quad 12 \ci 45, \quad 123 \ci 5.
$$
\end{example}
The stronger condition required by Definition~\ref{def:dbgm}
implies that in some situations not all marginal independence relations are representable
by bi-directed graphs, as the following example shows.
\begin{example}
Consider the data in Table~\ref{tab:lienert},
due to \citet{lienert:1970}.
The variables are 3 symptoms
after LSD intake, recorded to be present (level 1)
or absent(level 2), and are  distortions in affective behavior
($X_1$), distortions in thinking ($X_2$), and dimming of consciousness ($X_3$).
As \citet{wermuth:1998} points out,
the frequencies in the three marginal tables show that the
three symptom pairs are close to independence, but at the same time the variables are not
mutual independent as witnessed by the strong three-factor interaction due to
the quite distinct conditional odds ratios between $X_1$ and $X_2$ at the two levels of $X_3$.
Thus, in this case, despite three marginal independencies, a discrete bi-directed graph
model can represent just one of them, and thus  must include at least two edges.
\end{example}
\begin{table}[b]
  \centering
 \caption{\it Data by \citet{lienert:1970} concerning symptoms after LSD-intake.
$OR$ is the conditional odds-ratio between $X_1$ and $X_2$ given $X_3$. The frequencies
show evidence of pairwise independence, but mutual dependence.}
\label{tab:lienert}
  \begin{tabular}{lcrrrr}\hline
                     &$X_3$ & 1      &      & 2      &        \\
 $X_1$                 &$X_2$ & 1      &   2  & 1      &      2   \\ \hline
1                  &    & 21      &     5      &    4      &   16\\
2                   &    &  2      &     13      &    11      &   1\\ \hline
$OR$          &    & $27.3$         &    & $0.023$                 &   \\
  \end{tabular}
\end{table}
\citet{pea-wer:1994} studied the Markov equivalence between
bi-directed graph models (actually the covariance graphs) and
directed acyclic graphs models, i.e. when the two models imply exactly the
same conditional independence statements, under their respective global Markov property (for the global Markov property
see \citealp{lau:1996}). They showed that
each bi-directed graph is always Markov equivalent to a directed
acyclic graph with additional synthetic latent nodes, after marginalizing over the additional nodes, as exemplified  in
Figure~\ref{fig.introd}(b, c). Moreover they also give a Markov equivalence result,
proving that a bi-directed graph is equivalent to a directed acyclic
graph with the same set of nodes if and only if it contains no  4-chain. Thus, there is no
directed acyclic graph which is Markov equivalent to the bi-directed graphs of Figures~\ref{fig.introd}(a), \ref{fig.sec2}(a) or
\ref{fig.sec2}(b).
%
\section{Marginal log-linear parameterizations} \label{sec.mar}
Discrete bi-directed graph models may be defined
as marginal log-linear models, using complete hierarchical parameterizations as defined  by \citet{ber-rud:2002}.
In this section we review the basic concepts and we discuss the definitions of the parameters involved.
Let $p(\b i)>0$ be  a strictly positive probability distribution of a discrete random vector  $X = (X_v, v \in V)$
and let  $p_M(\b i_M)$  be any marginal probability distribution of a sub-vector $X_M$, $M \subseteq V$.
The marginal probability distribution admits a log-linear expansion
$$
\log p_M(\b i_M) = \sum_{L \subseteq M} \lambda^M_L(\b i_L)
$$
where $\lambda^M_L(\b i_L)$ is a function defining the log-linear
parameters indexed by the subset $L$ of $M$. The functions
$\lambda^M_L(\b i_L)$ are defined by
$$
\lambda^M_L(\b i_L) = \sum_{A \subseteq L} (-1)^{|L \setminus A|}
\log p_M(\b i_A, \b i^*_{M \setminus A})
$$
where $\b i^* = (1, \dots, 1)$ denotes a baseline   cell of the
table; see \cite{whi:1990} and \cite{lau:1996}. The function
$\lambda^M_L(\b i_L)$ is zero whenever at least one index in $\b
i_L$ is equal to 1. Therefore, $\lambda^M_L(\b i_L)$ defines  only
$\prod_{v \in L} (b_v-1)$ parameters where $b_v$ is the number of
categories of variable $X_v$. Due to the constraint on the
probabilities, that must sum  to one, the parameter
$\lambda^M_\phi = \log p(\b i^*_M)$  is a function of the others,
and can thus be eliminated.

If   $\b \lambda^M_L$ is the vector containing  the parameters $\b
\lambda^M_L(\b i_L)$, then it can be obtained explicitly using
Kronecker products as follows. For any subset $L$ of $M$, let $\b
C_{v,L}$ be the matrix
$$
 \b C_{v,L} =
\begin{cases}
 \left(-\b 1_{b_v-1} \quad \b I_{b_v-1}\right)  & \text{ if } v \in L \\
 \left( 1 \quad  \b 0_{b_v-1} \right)  & \text{ if } v \not \in L.\\
 \end{cases}
$$
and let $\b \pi^M$ be the $t_M\times 1$ column vector of the
marginal cell probabilities in lexicographic order. Then, the
vector of the log-linear parameters $\b \lambda^M_L(\b i_L)$ is
\begin{equation}\label{eq.c-matrix}
\b \lambda^{M}_{L} = \b C_{L}^{M} \log \b \pi^M, \text{ where } \b C^M_L =
\bigotimes_{v \in M} \b C_{v,L}.
\end{equation}
For a discussion of the technique of building all log-linear parameters based on Kronecker products  see
\citet{wer-cox:1992}. The coding used in this paper
corresponds to their indicator coding, and gives
the parameters used for example by the program \textsc{glim}.

A marginal log-linear parameterization of the probability distribution $p(\b i)$  is obtained by
combining the log-linear parameters $\b \lambda^M_L$ for many different  marginal probability distributions.
The general theory is developed in \cite{ber-rud:2002} and is summarized below.
\begin{defn}
\label{def:marloglin}
Let $\mcal = (M_1, \dots,  M_s)$ be an ordered sequence of margins of interest,
and, for each $M_j$, $j = 1, \dots, s$,
let $\lcal_j$ be the collection of sets $L$ for which $\b \lambda^{M_j}_L$ is defined
with equation \eqref{eq.c-matrix}. Then, $(\b \lambda^{M_j}_L)$
is said to be a hierarchical and complete marginal log-linear parameterization for $p(\b i)$ if
$(i)$ the sequence $M_1, \dots, M_s$ is non-decreasing;  $(ii)$ the last margin is $M_s = V$;
 $(iii)$ the sets defining the log-linear parameters in each margin are:
$$
\lcal_1 = \pcal_0(M_1), \text{ and } \lcal_j =
\pcal_0(M_j) \setminus \bigcup_{h = 1}^{j-1} \lcal_h, \text{ for } j > 1,
$$
where $\pcal_0(M_j)$ denotes the collection of all non-empty sets of $M_j$.
\end{defn}
The parameterization  is called hierarchical because it is
generated by a non-decreasing sequence $\mcal$, and complete because
it defines all possible log-linear parameters terms, each within one and only one
marginal table. Notice that the parameterization
is associated uniquely to a particular sequence $\mcal$ of margins. Thus,
a different (still non-decreasing) ordering of the sequence  induces
a different parameterization; see the examples in Section~\ref{sec.par.2}.

The above construction defines a map from the simplex $\Delta_{V}$
of the strictly
positive distributions $p(\b i)$ of the discrete
random vector $X$ into the set $\Lambda$  of possible
values for  the whole vector of the marginal log-linear parameters
$\b \lambda = (\b \lambda^{M_j}_L)$, with $j = 1, \dots, s$ and $L \in \lcal_j$.
The following  general result shows  that a complete hierarchical marginal log-linear model
defines a proper parameterization.
\begin{prop}\citep{ber-rud:2002}\label{teor:br}
The map $\Delta_V \rightarrow \Lambda \subseteq \mathbf{R}^{t-1}$ defined by
a complete and hierarchical marginal log-linear parameterization is a diffeomorphism.
\end{prop}
The  parameters $\b \lambda$ can be  written in matrix form
$$
 \b \lambda = \b C \log(\b T  \b\pi )
$$
where $\b \pi$ is the $t \times 1$ vector of all the cell probabilities in lexicographical order,
 $\b T$ is a $m \times t$ marginalization matrix such that
$$
\b T \b \pi = \begin{pmatrix}
\b \pi^{M_1} \\ \vdots \\ \b \pi^{M_s} \\
\end{pmatrix}
$$
and  $\b C = \diag(\b C^M_L)$ is a $t-1 \times m$ block diagonal matrix, with $m = \sum_{j=1}^s |\mathcal{I}_{M_j}|$.
For a discussion of algorithms for computing the matrices $\b C$ and $\b T$
see \citet{bar-col-for:2007}, that generalize the approach by \cite{ber-rud:2002}
to logits and higher order effects of global and continuation type, suitable with ordinal data .

The  log-linear  parameterization and the multivariate logistic
transformation represent two special cases of marginal log-linear
models. The standard log-linear  parameters are generated by
$\mcal = \{V\}$. They will be denoted by $\b \theta_L = \b
\lambda^V_L$ for $L \in \pcal_0(V)$ and the whole vector of
parameters by $\b \theta$. The parameter space coincides with
$\mathbf{R}^{t-1}$ and  the map from $\b \pi$ to $\b \theta$
admits an inverse in closed form, provided that $\b \pi > 0$. The
multivariate logistic parameters \cite{glo-mcc:1995} are generated
by $\mcal = \pcal_0(V)$, in any non-decreasing order. They will be
denoted  by $\b \eta^M = \b \lambda^M_M$, with $\b \eta$
representing the whole vector. Thus the parameters $\b \eta^M$
correspond to the highest order log-linear parameters within each
marginal table $\mathcal{I}_M$, for each nonempty set $M \subseteq
V$. The parameter space is in general a strict subset of
$\mathbf{R}^{t-1}$, except when the number of variables is $d =
2$. In general there is no closed form inverse transforming back
$\b \eta$ into $\b \pi$. The inverse operation however may be
accomplished using for example the iterative proportional fitting
algorithm.

Thus, while the log-linear parameters $\b \theta$ are always variation
independent and  for any  $\b \theta$ in $\mathbf{R}^{t-1}$ there is a  unique
associated joint probability distribution $\b \pi$,  instead the multivariate
logistic parameters  are never variation independent, for $d > 2$.
Thus there are vectors $\b \eta$ in  $\mathbf{R}^{t-1}$ that are not compatible
with any joint probability distribution $\b \pi$.
The latter assertion is also implied by a further result by
\citet{ber-rud:2002} which proves that the hierarchical and
complete marginal log-linear parameterization generated by a
sequence $\mcal$ is variation independent if and only if $\mcal$
satisfies a property called \emph{ordered decomposability}. A
sequence of arbitrary subsets of $V$ is said to be
ordered decomposable
if it has at most two elements or if there is an ordering $M_1, \dots, M_s$ of
its elements, such that $M_i \not\subseteq M_j $ if $i > j $  and, for $k = 3, \dots, s$, the maximal elements
(i.e. those not contained in any other sets) of $\{M_1, \dots,M_k\}$ form a decomposable set.
For further details and examples about ordered decomposability see \citet{ber-rud:2004}.
More properties of the two parameterizations $\b \theta$ and $\b \eta$, connected to graphical models,
 will described in the next Section~\ref{sec.par}.
\section{Parameterizations of discrete bi-directed graph models}\label{sec.par}
We   suggest now two different marginal log-linear
parameterizations of discrete bi-directed graph models, and we compare advantages and shortcomings.
\subsection{Multivariate logistic  parameterization} \label{sec.par.1}
It is known that the complete independence of  two
sub-vectors  $X_A, X_B$ of the random vector $X$   is equivalent to a set of zero restrictions on multivariate logistic parameters.
\begin{lemma} {\rm (\citet{kau:1997}, Lemma 1)}.\label{lemma1}
If $\{A, B\}$ is a partition of $V$ and  $\b \eta = (\b \eta^M), M \in \pcal_0(V)$ is the multivariate
logistic parameterization,   then
\begin{equation}\nonumber
A \ci B \iff  \quad \b \eta^{M} = \b 0 \quad \text{ for all  } M \in \qcal
\end{equation}
where $\qcal = \{ M \subseteq A \cup B:  M\cap A \ne  \emptyset, M \cap B \ne \emptyset\}$.
\end{lemma}
We generalize  this result to complete independence of more than two random vectors.
Given a partition $\{C_1, \dots, C_r\}$ of a set $D \subseteq V$,  we define
$$
\qcal(C_1,\dots,C_r) = {\textstyle \pcal\left(\bigcup_{i=1}^r C_k\right) \setminus  \bigcup_{i=1}^r}
\pcal(C_k).
$$
This is the set of all subsets of $D$ not completely contained in a single class, i.e.
containing elements coming from at least two classes of the partition. With  this notation,
the set $\qcal$ of Lemma~\ref{lemma1}  may be denoted by $\qcal(A,B)$. Then we have the following result.
\begin{prop}\label{pro.indip-gener}
Let $X=(X_v), v\in V$,  be the discrete random  vector with multivariate
logistic parameterization  $\b \eta = (\b \eta^M), M \in \pcal_0(V)$.
If $D \subseteq V$ is partitioned into the classes $\{C_1, \dots, C_r\}$ then
$$
C_1 \ci \dots \ci C_r \iff  \text{ for all } M \in
\qcal(C_1,\dots,C_r ):\quad \b \eta^{M} = \b 0.
$$
\end{prop}
\begin{proof}
First, use the shorthand notations $\qcal$ to denote the set $\qcal(C_1,\dots,C_r)$ and
$\qcal_i$ to denote the set $\qcal(C_i,C_{-i})$, $i =1,\dots,r$,
where $C_{-i} = D \setminus C_i$. In fact,
since $\qcal_i \subseteq \qcal$, then $\bigcup_{i=1}^r{\qcal}_i\subseteq{\qcal}$.
Conversely, for any  $M\in{\qcal}$ there is always  a class $C_i$ such that $C_i \varsubsetneq M$, and
hence, by definition,  $M \in {\qcal}_i$. Hence, for every $M \in \qcal$,  $M \in \bigcup_{i=1}^r{\qcal}_i$ and thus ${\qcal}\subseteq \bigcup_{i=1}^r{\qcal}_i$.
Then,  the complete  independence $C_1 \ci \cdots \ci C_r$ is equivalent
to $C_i \ci C_{-i}$ for all $i = 1, \dots, r$.
By  Lemma~\ref{lemma1}, applied to the sub-vector $X_D$,
each independence $C_i \ci C_{-i}$ is equivalent to the restriction $\b \eta^M = \b 0$ for
$M \in \qcal_i$ and the parameters $\b \eta^M$ are identical to the corresponding
multivariate logistic parameters for the full random vector $X_V$.
Thus,  the complete independence $C_1 \ci \cdots \ci C_r$
is equivalent to  $\b \eta^M=\b 0$ for
$M \in {\qcal}_i$, $i = 1, \dots, r$, i.e. for $M \in \bigcup_{i=1}^r{\qcal}_i = \qcal$.
\end{proof}
Proposition~\ref{pro.indip-gener} implies that a statement of
complete independence $C_1 \ci \dots \ci C_r$ is equivalent to
a set of zero constraints on the multivariate logistic  parameters. The
following result explains how the constraints must be chosen in order to
satisfy all the independencies required by the Definition~\ref{def:dbgm} of a bi-directed graph model.
\begin{prop}
\label{prop.cov.m-logit}
Given a bi-directed graph $G = (V, E)$, the discrete bi-directed graph model associated with  $G$
is defined by the set of strictly positive discrete probability distributions with multivariate logistic parameters
$\b \eta =(\b \eta^M)$,   $M \in \pcal_0(V)$, such that
\begin{equation}
\b \eta^{M} = \b 0 \text{ for every }  M \in \dcal, \nonumber
\end{equation}
where $\dcal$ is the set of all  disconnected sets of nodes in the graph $G$.
\end{prop}
\begin{proof}
Given a set $D \in \dcal$, denote its  connected components by $\{C_1, \dots C_r\}$ and by
$\qcal_D$ the set  $\qcal(C_1, \dots, C_r)$.
First,  we prove that $\dcal =\bigcup_{D \in \dcal} \qcal_D$.  In fact, for any $D\in\dcal$,
$\qcal_D \subseteq \dcal$ because it
is  a class of disconnected
subsets of $D$. Thus, $\bigcup_{D \in \dcal} \qcal_D \subseteq \dcal$.
Conversely, if $D \in \dcal$, then $D \in \qcal_D$ and thus
$\dcal \subseteq \bigcup_{D \in \dcal} \qcal_D$.
By Definition~\ref{def:dbgm},  the independence
$C_1 \ci \cdots \ci C_r$ is implied  for each disconnected set $D$ with connected components $C_1, \dots, C_r$.
By Proposition~\ref{pro.indip-gener}, this  is equivalent to the zero restrictions on the multivariate logistic
parameters
$$
\b \eta^M = \b 0, \text{ for all } M \in \qcal_D, \quad D \in \dcal
$$
i.e. for all $M \in \bigcup_{D \in \dcal} \qcal_D = \dcal$.
\end{proof}
\begin{table}
\centering
\caption{\it Comparison between  two
parameterization of the discrete chordless 4-chain model of Figure~\ref{fig.introd}(a): ($\b \eta$) with bi-directed edges;  ($\b \theta$) with undirected edges.}
\begin{small}
\begin{tabular}{lccccccccccccccc}
 \hline
Terms      & 1         & 2        & 3        & 4        & 12           & 13
& 14 & 23          & 24 & 34 & 123    & 124 & 134   & 234 & 1234 \\ \hline
$\be$     & $\be^1$  & $\be^2$   & $\be^3$   & $\be^4$   & $\be^{12}$
& $\b 0$  & $\b 0$  & $\be^{23}$    &  $\b 0$  & $\be^{34}$    & $\be^{123}$ & $\b 0$     & $\b 0$
& $\be^{234}$ & $\be^{1234}$ \\
$\bt$   & $\bt_1$  & $\bt_2$ & $\bt_3$ & $\bt_4$ &
$\bt_{12}$  & $\b 0$    & $\b 0$    & $\bt_{23}$ & $\bf 0$    & $\bt_{34}$  & $\bf 0$
& $\b 0$     & $\b 0$    & $\b 0$   & $\b 0$    \\ \hline
\end{tabular}
\end{small}\label{tab:compar}
\end{table}
A consequence of Proposition~\ref{prop.cov.m-logit} is that all possible discrete
bi-directed graphical models can be identified within the multivariate logistic parametrization under the
zero constraints associated with the disconnected sets.
\bex
The discrete model  associated with the chordless 4-chain of Figure~\ref{fig.introd}(a)
is defined by the multivariate logistic parameters shown in Table~\ref{tab:compar},
first row. There are 5  zero constraints  on the highest-order log-linear
parameters of the tables 13, 14, 24, 124 134.
There  are three nonzero two-factor marginal log-linear
parameter $\be^{ij}$ associated with the edges of the graph
that  may be interpreted as sets of marginal
association coefficients between the involved variables,
based on the chosen contrasts.
Consider now the reduced model resulting after dropping the edge $2 \leftrightarrow 3$ and implying the independence
$12 \ci 34$. This model can be obtained, within the same parameterization,
by the additional zero constraints
on  $\be^{23}, \be^{123}, \be^{234}$ and $\be^{1234}$.
\eex
While the parameters are in general  not variation independent, they satisfy
the upward compatibility property, because  they
have the same meaning  across different marginal distributions.
Using this property, we can  prove the following result
concerning the effect of marginalization over a subset $A$ of the variables.
Let $G_A = (A, E_A)$ be the subgraph induced by $A$, and
let $\dcal_A$ be  the set of all disconnected sets of $G_A$.
\begin{prop}
If a discrete probability distribution $p(\b i)$ for $\b i \in \mathcal{I}_V$ satisfies a
bi-directed graph model  defined by the graph $G =(V, E)$
then the marginal distribution $p_A(\b i_A)$ over $A \subseteq V$
satisfies the bi-directed graph model defined by $G_A = (A, E_A)$ and
its  multivariate logistic parameters  are $\b \eta = (\b \eta^M), M \in \pcal_0(A)$
with constraints $\b \eta^M = \b 0$, for $M \in \dcal_A$.
\end{prop}
\begin{proof}
After marginalization over $A$, the multivariate logistic parameters associated with $p_A(\b i_A)$,
by the property of upward compatibility,  are   $(\b \eta^M, M \in \pcal_0(A))$.
Some of these parameters are zero by the  constraints implied by the
 original bi-directed graph model, i.e.  $\b \eta^M = \b 0$,
for $M \in \dcal \cap \pcal_0(A)$.
The result is proved by showing that  $\dcal \cap \pcal_0(A) = \dcal_A$.
First, we note that if $D \subseteq A \subseteq V$, then
the graph $G_D = (D, E_D)$ with edges $E_D = (D \times D) \cap  E = (D \times D) \cap E_A$ is a subgraph of both $G_A$ and $G$.
Thus, if $D \subseteq A$ and $D \in \dcal$ then the induced subgraph $G_D$ is disconnected and being also  a subgraph of $G_A$
then $D$ is also a disconnected set of $G_A$. Thus $\dcal \cap \pcal_0(A) \subseteq \dcal_A$.
Conversely, if $D$ is a disconnected set of $G_A$,
then the subgraph $G_D$ is disconnected, and being a subgraph of $G$, then
$D$  is also a disconnected set of $G$.
Thus $\dcal_A \subseteq \dcal \cap \pcal(A)$, and the result follows.
\end{proof}
Discrete bi-directed graph models in the multivariate logistic parameterization can be compared
with discrete log-linear graphical models represented by undirected graphs with the same skeleton
(i.e. with the same set $E$).
To facilitate the comparison we state the following well-known result, following from the Hammersley and Clifford
 theorem, \citep[see][p.~36]{lau:1996},  which is the undirected graph model counterpart of   Proposition~\ref{prop.cov.m-logit}.
\begin{prop}
\label{prop.con.loglin}
Given an undirected  graph $G=(V,E)$,  a  discrete graphical log-linear model associated with $G$ is defined by the set of strictly positive
discrete probability distributions with log-linear parameters
$\b \theta = (\b \theta_L, L \in \pcal_0(V))$, such that
\begin{equation}
\b \theta_{L} = \b 0 \text{ for every }  L \in \mathcal{N}, \nonumber
\end{equation}
where $\mathcal{N}$ is the set of all  incomplete subsets of nodes in the
graph $G$.
\end{prop}
The set $\dcal$ of all disconnected sets of a graph $G$
 is  included in the set $\mathcal{N}$ of the incomplete sets, and therefore
the number of zero restrictions of the undirected graph models is always higher than the number of zero restrictions
of the bi-directed graph models with the same skeleton, (see \citealp{drt-ric:2007}).
\begin{example}
A discrete undirected graph model for the 4-chain implies the independencies $ 12 \ci 4 | 3$ and $ 1 \ci 34 | 2$ and is defined by zero
constraints on 8 log-linear parameters $\b \theta_L$,  shown in  Table~\ref{tab:compar}, second row.
Also, Proposition~\ref{prop.con.loglin} implies
that in the discrete  undirected graph model  the  general hierarchy principle holds, i.e.
if a particular log-linear  term  is zero then all higher terms containing the same set of subscripts are
also set to zero. On the contrary, by Proposition~\ref{prop.cov.m-logit},
in the multivariate logistic parameterization of the bi-directed graph model
the hierarchy principle is violated because a superset of a disconnected set may be connected.
Thus, for instance in the example shown in Table~\ref{tab:compar} there are zero pairwise associations,
like $\be^{13} = \b 0$, but nonzero higher order log-linear parameters like
$\be^{123} \ne \b 0$ and $\be^{1234} \ne \b 0$.
\end{example}


\subsection{The disconnected sets parameterization}\label{sec.par.2}
We discuss now another marginal log-linear parameterization that can represent the
independence constraints implied by any discrete bi-directed graph model, but
involving only those marginal tables which are needed. This parameterization defines
the log-linear parameters  within the margins associated with the disconnected sets of the
graph defining the model. Specifically, given a discrete graph model with a graph $G$,
we  arbitrarily order the disconnected sets of the graph to yield  a non-decreasing sequence
$(D_1,\dots,D_s)$ such that $D_{k}
\not \supseteq D_{k+1}$ for $k = 1, \dots, s-1$. Then, the  \emph{disconnected set parameterization}
of the discrete bi-directed graph model associated with $G$, is the hierarchical and complete marginal
log-linear parameterization $\b \lambda = (\b \lambda^{M_j}_L)$ generated, following Definition~\ref{def:marloglin}, by the sequence of  margins
\begin{equation}
\mcal_G = \begin{cases}
(D_1,\dots,D_s)  & \text{ if }   D_s = V\\
(D_1,\dots,D_s, V)& \text{otherwise.}\\
\end{cases}
\label{eq.mg}
\end{equation}
This parameterization contains by definition the log-linear  parameters $\b \lambda^D_D = \b \eta^D$ for every disconnected
set $D$ and thus can define the independence model by the same constraints of Proposition~\ref{prop.cov.m-logit}.
\begin{prop}\label{pro.MLL-COV}
Given a bi-directed graph $G = (V, E)$, the discrete bi-directed graph model
associated with  $G$ is defined by the set of strictly positive discrete
probability distributions with a disconnected set parameterization
$(\b \lambda^{M_j}_L)$, such that
$$
\b \lambda^{M_j}_{M_j} = \b 0 \text{ for every } M_j \in \dcal,
$$
where $\dcal$ is the class of all disconnected sets for $G$.
Moreover, the constraints are independent of the ordering chosen
to define $\mcal_G$.
\end{prop}
\begin{proof}
The disconnected set parameterization defined by the sequence
\eqref{eq.mg}, contains the parameters $\b \lambda^D_L$, with $D
\in \dcal$. By Definition~\ref{def:marloglin}, $\lcal_j$, $j = 1,
\dots, s$ always contains the set $D$ itself. This happens
whatever ordering is used to define $\mcal_G$. Thus the
parameterization always includes $\b \lambda^{D}_{D} = \b \eta^D$,
for every $D \in \dcal$ and it is possible to impose the
constraints $\b \eta^{D} = \b 0$ for every $D \in \dcal$ and  the
result follows by Proposition~\ref{prop.cov.m-logit}.
\end{proof}
While the constrained parameters defining the bi-directed graph model are actually
the same as the multivariate logistic parameterization, the other unconstrained
log-linear parameters are  defined in larger  marginal tables, and thus have a different
interpretation. An important difference is that
the disconnected set parameterization is tied to the specific
graph $G$ defining the model. This implies that it is not
possible to define every bi-directed graph model within the same disconnected
set parameterization.
A different model $G$ implies a different sequence $\mcal_G$ of disconnected sets
and thus a different list of log-linear parameters.
\bex
For the chordless 4-chain graph of  Figure~\ref{fig.introd}(a),
there are several possible orderings of the 5 disconnected sets
$\dcal = \{13, 14, 24, 134, 124\}$. The discrete bi-directed graph model is defined
by choosing for example
$$
\mcal_G = (13, 14, 24, 134, 124, 1234),
$$
and by constraining the marginal log-linear
parameters $\b \lambda^{D}_{D} = \b  0$ for $D \in \dcal$.
The unconstrained parameters differ
from the multivariate logistic ones.
 For example the two-factor log-linear parameters
between $X_1$ and $X_2$,
$\b \lambda^{124}_{12}$, are defined within the marginal table 124
instead of the marginal table 12.
A detailed comparison
between the parameters is reported in the first two rows of the
Table~\ref{tab:compar}.
\eex
\begin{table}[t]
\begin{center}
\caption{\it Comparison of three parameterizations for the bi-directed graph
model $G$ of Figure~\ref{fig.introd}(a). One-factor log-linear parameters are omitted. The columns
of parameters  to be constrained to zero have a boldfaced label.}
\label{tab:compar2}
\small
\renewcommand{\arraystretch}{1.2}
\begin{tabular}{lccccccccccc}
 \hline
Terms      & 12           & {\bf 13} & \textbf{14} & 23        & \textbf{24} & 34 & 123    & \textbf{124} & \textbf{134}   & 234 & 1234 \\ \hline
$\be$     & $\be^{12}$
& ${\be^{13}}$  &  ${\be^{14}}$  & $\be^{23}$    &  ${\be^{24}}$   & $\be^{34}$    & $\be^{123}$ &   ${\be^{124}}$ &   ${\be^{134}}$
& $\be^{234}$ & $\be^{1234}$ \\
$\mcal_G$     &  $\bl^{124}_{12}$    & ${\bl^{13}_{13}}$  &  ${\bl^{14}_{14}}$   &
$\bl^{1234}_{23}$    & ${\bl^{24}_{24}}$   & $\bl^{134}_{34}$    &
$\bl^{1234}_{123}$ &  ${\bl^{124}_{124}}$   &  ${\bl^{134}_{134}}$   & $\bl^{1234}_{234}$ &
$\bl^{1234}_{1234}$ \\
$\mcal'_G$     & $\bl^{124}_{12}$    & ${\bl^{134}_{13}}$   &  ${\bl^{14}_{14}}$   &
$\bl^{1234}_{23}$    & ${\bl^{124}_{24}}$   & $\bl^{134}_{34}$    &
$\bl^{1234}_{123}$ &  ${\bl^{124}_{124}}$   & ${\bl^{134}_{134}}$     & $\bl^{1234}_{234}$ &
$\bl^{1234}_{1234}$\\
 \hline
\end{tabular}
\end{center}
\end{table}
The previous example shows that  we can collect the log-linear
parameters into a reduced number of marginal tables. An
alternative selection of marginal tables could be chosen in order
to fulfill the conditional independencies implied by the global
Markov property. We will describe  the method in the special case
of the chordless 4-chain graph. It is conjectured that a general
variation independent parameterization does not exists for all
bi-directed graphs, but the definition of a sub-class admitting
such a parameterization is still an open problem. \bex \label{sur}
 In Example~\ref{ex.glob} we stated that, for the
bi-directed 4-chain graph of Figure~\ref{fig.introd}(a),
the global Markov property implies  the conditional
independencies  $1 \ci 4$, $ 2\ci 4| 1$ and $1 \ci 3|4$.
Thus, the relevant margins can be  collected in the sequence
$$
\mcal'_G = (14, 134, 124, 1234)
$$
where the first three allow the definition of the conditional independencies and the last one
serves as  completion of the parameterization.  The complete hierarchical parameterization
generated by $\mcal'_G$ is slightly different from that generated by $\mcal_G$, see
 Table~\ref{tab:compar2}, third row, but with the 5 zero constraints on the higher
level log-linear parameters within each margin, we obtain the required independencies
$$
1 \ci 4 \iff  \b \lambda^{14}_{14} = \b 0 \quad
2 \ci 4 | 1  \iff
 \begin{cases}
\b \lambda^{124}_{24}   = \b 0 & \\
  \b \lambda^{124}_{124}  = \b 0 &  \\
\end{cases}\quad
1 \ci 3 | 4  \iff
\begin{cases}
\b \lambda^{134}_{13}  = \b 0 & \\
\b \lambda^{134}_{134} = \b 0. & \\
\end{cases}
$$
Note that these  independencies can also be represented  by  a chain graph with two components,
$\{1,4\}$ and  $\{2,3\}$, under the alternative Markov property, (see
\citealp{amp:2001}). The associated discrete model is interpreted
as  a system of seemingly unrelated regressions, with two joint responses
$X_2$ and $X_3$.  In this context the associations of interest are
the effect parameters  between every response and each explanatory variable
conditional on the remaining explanatory variable, i.e.
$\b \lambda_{12}^{124}$, $\b \lambda_{24}^{124}$, $\b \lambda_{13}^{134}$ and $\b \lambda_{34}^{134}$, and
the marginal association parameters between the explanatory variables, $\b \lambda_{14}^{14}$.
By relaxing the constraint $\b \lambda^{14}_{14} = \b 0$ we obtain
a discrete chain graph model with  two complete chain components, under the alternative
Markov property.
\eex
In the comparison between different parameterizations also the property of variation
independence may be relevant.  Following \citet{ber-rud:2002},
given a discrete bi-directed graph model, there is a variation
independent parameterization if there is at least a sequence  $\mcal_G$
which is ordered decomposable. This property  is quite relevant because
the lack of variation independence may make the separate interpretation of
the parameters misleading.
\bex
\label{ex.house}
In the previous example both the parameterizations based on $\mcal_G$ and $\mcal'_G$ are
variation independent (unlike the multivariate logistic parameterization)
because the sequences of margins are both ordered decomposable.
Consider instead the bi-directed graph in Figure \ref{fig.sec2}(a). Two possible
disconnected set parameterizations of the discrete model may be based for example on
\begin{eqnarray*}
\mcal_G &=& (13, 14, 25, 35, 134, 135, 235, 12345), \\
\mcal_{G}' &=& (13, 35, 135, 14, 25, 134, 235, 12345).
\end{eqnarray*}
with the constraints $\b \lambda^D_D = \b 0$ for any disconnected set $D$.
In this case we can verify that only the sequence $\mcal'_G$  is ordered decomposable
and thus implies variation independent parameters.
\eex

\section{Maximum likelihood estimation of discrete bi-directed graph models}\label{sec.fit}
We study now the maximum likelihood estimation of the discrete bi-directed graph models
under any of the parameterizations previously discussed.
Assuming a multinomial sampling
scheme with sample size $N$, each individual falls in a cell $\b i$ of
the given contingency table
$\mathcal{I}_V$  with probability $p(\b i)>0$. Let $n(\b i)$ be the
cell count and $\b n = (n(\b i), \b i \in \mathcal{I}_V)$, be a $t \times 1$ vector. Thus,
$\b n$ has a multinomial distribution with parameters $N$ and $\b \pi$. If
 $\b \mu = N \b \pi > \b 0$ is the expected value of $\b n$ and $\b \omega = \log \b \mu$,
then for any appropriate marginal log-linear parameterization $\b \lambda$
we have $\b \lambda = \b C \log (\b T \b \pi) = \b C \log (\b T \exp(\b \omega))$ because the contrasts
of marginal probabilities are equal to the contrasts of expected counts.
Given a discrete bi-directed graph model defined by the  graph $G = (V, E)$,
 if $\b \lambda$ is defined
either by the multivariate logistic parameterization or by the disconnected
set parameterization,
we can always split  $\b \lambda$ in two components  $\b \lambda_{\dcal}$ and
$\b \lambda_{\ccal}$ indexed
by the disconnected sets $\dcal$  and by the connected sets $\ccal$ of the graph, respectively.
If $\b C_{\dcal}$ is a sub-matrix of  the contrast matrix $\b C$,
obtained by selecting the rows
associated with the disconnected sets of the graph $G$,
$$
\b \lambda_{\dcal} =  \b C_\dcal \log (\b T \exp(\b \omega)) = \b h(\b \omega)
$$
where  $\b C_\dcal$ has dimensions   $q \times v$   with
$q = \sum_{D \in \mathcal{D}} \prod_{v \in D} (b_v-1)$. Thus, the kernel of the
log-likelihood function
of the discrete bi-directed graph model is defined by
\begin{equation}\label{eq.lik}
l(\b   \omega; \b n) =  \b n\T \b \omega - \b 1\T \exp(\b \omega), \quad \b \omega \in \Omega_{BG},
\end{equation}
with
$$
\Omega_{BG} = \{\b \omega \in \mathbf{R}^t: \b h(\b \omega) = \b 0,
\quad \b 1\T \exp(\b \omega) = N\}.
$$
Note that \eqref{eq.lik} defines a curved exponential family model
as the set $\Omega_{BG}$ is a smooth manifold
in the space $\mathbf{R}^t$ of the canonical parameters $\b \mu$.  Maximum likelihood estimation
is a constrained optimization problem and the
 maximum likelihood estimate is  a saddle
point of the Lagrangian log-likelihood
$$
\ell(\b \omega, \b \tau) = \b n\T \b \omega - \b 1\T \exp(\b \omega) +
\b \tau\T \b h(\b \omega)
$$
where $\b \tau$ is a $q \times 1$ vector of unknown Lagrange multipliers.
To solve the equations we propose an
iterative procedure inspired by \citet{ait-sil:1958}, \citet{lang:1996} and
\citet{ber:1997}.
Define first
$$
\b \xi = \begin{pmatrix}
\b \omega \\
\b \tau
\end{pmatrix}, \quad
\b f (\b \xi) = \frac{\partial \ell}{\partial \b \xi}
=\begin{pmatrix}
\b f_\omega \\
\b f_\tau \\
\end{pmatrix}  \quad
\b F (\b \xi) =- E\left( \frac{\partial^2 \ell}{\partial \b \xi \partial \b \xi\T}\right) =
\begin{pmatrix}
\b F_{\omega\omega} & \b F_{\omega\tau}\\
\cdot & \b F_{\tau\tau}\\
\end{pmatrix},
$$
where the dot  is a shortcut to denote a symmetric sub-matrix.
Differentiating the Lagrangian  with respect to $\b \omega$ and $\b \tau$ and equating the result to zero
we obtain
\begin{equation}\label{eq-lagrange}
\begin{pmatrix}
\b f_\omega \\ \b f_\tau \\
\end{pmatrix} = \begin{pmatrix}
\b e + \b H\b \tau \\
\b h(\b \omega) \\
\end{pmatrix} = \b 0
\end{equation}
where
$\b e = \partial l/\partial \b \omega = \b n - \b \mu $,  $ \b H = \partial \b h/\partial \b \omega\T =  \b D_{\mu} \b T\T \b D^{-1}_{T\mu} \b C\T_\dcal$
and $\b D_{T\mu}$ and $\b D_\mu$ are diagonal matrices, with nonzero
elements $\b T\b \mu$ and $\b \mu$, respectively.

Let $\hat{\b \omega}$ be a local maximum of the likelihood subject
to the constraint $\b h(\b \omega) = \b 0$. A classical result
\citep{bert:1982} is that if $\b H$ is of full column rank at
$\hat{\b \omega}$, there is a unique $\hat{\b \tau}$ such that
$\ell(\hat{\b \omega},\hat{\b \tau}) = \b 0$. In the sequel, it is
assumed that the maximum likelihood estimate $\hat{\b \omega}$ is
a solution to the equation \eqref{eq-lagrange}. Note that the
constraint $\b 1\T \b \mu = \b 1\T \b n$ is automatically
satisfied as it can be verified that $\b H\T \b 1 = \b 0$ and thus
from \eqref{eq-lagrange} it follows that  $\b 1\T \b e = \b 0$.

Aitchison and Silvey propose a Fisher score like updating function
\begin{equation}
\label{eq.update}
\b \xi^{(k+1)} = \b u(\b \xi^{(k)}), \text{ with }   \b u(\b \xi) = \b \xi + \b F^{-1}(\b \xi)  \b f(\b \xi),
\end{equation}
yielding the estimate $\b \xi^{(k+1)}$ at cycle $k+1$ from that at
cycle $k$. As the algorithm  does not always converge when
starting estimates are not close enough to $\hat{\b \omega}$, it
is necessary to introduce a step size into the updating equation.
The standard approach to choosing a step size in optimization
problems is to use a value for which the objective function to be
maximized increases. However, since in in this case we are looking
for a saddle point of the Lagrangian likelihood $\ell$, we need to
adjust the standard strategy. First, the matrix $\b F$ has a
special structure with $\b F_{\omega\omega} = \b D_\mu$, $\b
F_{\omega\tau} = -\b H$ and $\b F_{\tau\tau} = \b 0$. Thus,
indicating the sub-matrices of $\b F^{-1}$ by superscripts, we
have $\b F_{\tau\omega}\b F^{\omega\tau} = \b I$  and $\b
F^{\omega\omega} \b F_{\omega\tau} = \b 0$. Thus the updating
function $\b u(\b \xi)$ of \eqref{eq.update} can be rewritten  as
follows
$$
\b u_\omega(\b\omega) = \b \omega + \b F^{\omega\omega} \b e + \b F^{\omega\tau} h(\b \omega),\quad
\b u_\tau(\b \omega) =  \b F^{\tau\omega} \b e + \b F^{\tau\tau}h (\b \omega),
$$
neither of which is a function of $\b \tau$.
As the updating  of the Lagrange
multipliers does non depend on the estimation for $\b \tau$
at previous step, the algorithm essentially searches in
the space of $\b \omega$. Hence, inserting a step size is only
required for updating $\b \omega$ and
we propose, following \citet{ber:1997} to use the following basic updating equations  with an added
step size, $0 < \mathrm{step}^{(k)} \le 1$:
$$
 \b \omega^{(k+1)}  =
 \b \omega^{(k)} + \mathrm{step}^{(k)}  \{\b F^{\omega\omega(k)} \b e^{(k)} + \b F^{\omega\tau(k)} h(\b \omega^{(k)})\},
$$
where $\b e^{(k)} = \b n - \hat{\b \mu}^{(k)}$ and where
$\b F^{\omega\omega(k)}$ and $\b F^{\omega\tau(k)}$ are two sections of
 $\hat{\b F}^{-1}$ at cycle $k$.
We chose the step size by a simple  step halving criterion, but
more sophisticated step size rules could also be considered. A discussion on the choice of the step size may be found in
\citet{ber:1997}.
Note that the algorithm's updates take place  in the rectangular space $\mathbf{R}^t$ of $\b \omega$
rather than the not necessarily rectangular space $\Lambda$ of the marginal log-linear parameters which may not be variation independent.
The algorithm converges if it is started
from suitable initial estimates of $\b \omega$ and $\b \tau$.
While  usually  a zero vector is a good
choice for $\b \tau$, we found empirically that the number of iterations to convergence
can be reduced substantially by using as a starting value for $\b \omega$
an approximate maximum likelihood estimate based on results by \citet{cw:1990} and \citet{roddam:2004}.
At convergence, we obtain the maximum likelihood estimates $\hat{\b \mu} = \exp(\hat {\b \omega})$ and
$\hat{\b \pi} = N^{-1}\hat{\b \mu}$ and the asymptotic covariance matrices
$$
\mathrm{cov}(\hat{\b \omega}) = \hat{\b  F}^{\omega\omega}, \qquad
\mathrm{cov}(\hat{\b \lambda}) = \b H_{sat} \hat{\b F}^{\omega\omega} \b H_{sat}\T, \text{ with }
\b H_{sat} = \b D_{\hat \mu} \b T\T \b D^{-1}_{T\hat \mu} \b C\T.
$$

\section{Analysis of some examples}\label{sec.app}
The examples of this section illustrate both the parameterizations and the fitting
of  marginal independence models. It is rare that a pure marginal independence model
is useful in isolation and thus usually it is interpreted  in combination
with  other  graphical models. However, the problem of simultaneous testing of
multiple marginal independencies in a general contingency table is often present
in applications and it can be carried out with the technique discussed in this paper.
All the computations were programmed  in the \textsf{R} language \citep{erre}.
\begin{example}
\begin{table}
\centering
\caption{\it Parameters estimates of the 4-chain model for the data on
symptoms of psychiatric patients under the multivariate logistic
and the  disconnected set  parameterizations. The fit is $\chi^2_5 = 8.61$. Columns (1)
and (2) are studentized estimates.}\label{tab.psy}
\small
\begin{tabular}{lrrllrr}\hline
\multicolumn{3}{c}{Multivariate logistic param.} & \multicolumn{4}{c}{Disconnected set param.}\\
Margin  &   $\hat{\b \eta}$ & (1)&  Margin    & Interaction& $\hat {\b \lambda}$ & (2) \\ \hline
1       &   $-0.28$  &  $ -2.62$  &  13        & 1          &  $-0.28$  & $-2.62 $    \\
2       &   $-0.13$  &  $ -1.23$  &            & 3          &  $ 0.21$  & $ 1.95 $    \\
3       &   $ 0.21$  &  $  1.95$  &            & 13         &  $ 0.00$  &      \\ \cline{4-7}
4       &   $ 0.24$  &  $  2.31$  &  14        & 4          &  $ 0.24$  & $ 2.31 $    \\
12      &   $-0.72$  &  $  -3.47$ &            & 14         &  $ 0.00$  &             \\ \cline{4-7}
13      &   $ 0.00$  &             &  24        & 2          &  $-0.13$  & $-1.23 $    \\
14      &   $ 0.00$  &             &            & 24         &  $ 0.00$  &              \\ \cline{4-7}
23      &   $-1.12$  &  $ -5.32$  &  124       & 12         &  $-0.72$  & $-3.47 $    \\
24      &   $ 0.00$  &             &            & 124        &  $ 0.00$  &               \\ \cline{4-7}
34      &   $ 0.79$  &  $  3.80$  &  134       & 34         &  $ 0.79$  & $ 3.80 $    \\
123     &   $ 0.16$  &    $0.36$  &            & 134        &  $ 0.00$  &             \\ \cline{4-7}
124     &   $ 0.00$  &             &  1234      & 23         &  $-0.78$  & $-1.80 $    \\
134     &   $ 0.00$  &             &            & 123        &  $ 0.14$  & $ 0.20 $    \\
234     &   $-0.90$  &  $ -2.03$  &            & 234        &  $-1.02$  & $-1.63 $    \\
1234    &   $ 0.15$  &  $  0.16$  &            & 1234       &  $ 0.15$  & $ 0.16 $    \\ \hline
\end{tabular}
\end{table}
The 4-chain marginal independence model was fitted to the data
on symptoms of psychiatric patients of Table~\ref{tab.coppen} with the algorithm of
Section~\ref{sec.fit}. After 22 iterations, the algorithm leads to a chi-squared goodness of fit
of 8.61 on 5 degrees of freedom. By comparison, the  best graphical log-linear model
has generators $[12][234]$ with  a deviance of $8.4$ on 6 degrees of freedom.
Thus, both models provide adequate interesting interpretations of the data.
Table~\ref{tab.psy} summarizes the estimates of the 4-chain graph model,  showing the parameter estimates and the
studentized estimates under the multivariate logistic and the disconnected set
parameterizations. In the multivariate logistic parameterization the
two-factor parameters have the simple interpretation of marginal association coefficients.
It must be kept in mind that they measure
just the strength of marginal association between  pairs of adjacent variables
in the graph, but that the model includes higher order log-linear parameters which
are not visible from the graph.
For instance, both $\hat\eta^{23}=-1.12$ and $\hat \eta^{234}=-0.90$
are measures of association for variables $X_2$ and $X_3$. In general,
for any connected subgraph, all higher order log-linear parameters are expected.
As explained in Section~\ref{sec.par},
the interpretation of the parameters necessarily
depends on the chosen parameterization.
For instance,  $\hat \eta^{23} = -1.12$ and $\lambda^{1234}_{23} = -0.78$ are a marginal
association measure and a conditional association measure respectively.
 The four-factor log-linear parameter is not significant, and a simpler reduced model with the
additional zero constraint on this parameter, has an adequate  chi-squared goodness
of fit of $8.63$ on 6 degrees of freedom.
\end{example}
\begin{table}
\caption{\it Data from U.S. General Social Survey.} \label{tab.usgss}
\small
\centering
\begin{tabular}{cccccrrrrrrrrr} \hline
   &       &       &       & $F$       &   1&    &    &   2&    &    &   3&    &    \\
$S$  & $C$     & $G$     & $A$     & $J$       &   1&   2&   3&   1&   2&   3&   1&   2&   3 \\ \hline
m  & 1     & 1     & 1     &         & 410& 241&  80& 691& 556& 187& 192& 148&  84 \\
   &       &       & 2     &         &  71&  31&   9& 109&  64&  34&  27&  26&  15 \\
   &       & 2     & 1     &         & 181& 128&  42& 307& 284&  82&  84&  93&  41 \\
   &       &       & 2     &         &  41&  17&   5&  61&  35&  20&  18&  13&   5 \\
   & 2     & 1     & 1     &         &  96&  77&  29& 163& 151&  76&  58&  55&  27 \\
   &       &       & 2     &         &  34&  18&   7&  58&  36&  15&  17&  13&   6 \\
   &       & 2     & 1     &         &  29&  37&   4&  55&  54&  31&  22&  26&  17 \\
   &       &       & 2     &         &  16&   6&   6&  16&  16&   7&  10&   7&   2 \\
f  & 1     & 1     & 1     &         & 552& 353& 145& 899& 793& 265& 180& 162&  94 \\
   &       &       & 2     &         &  98&  60&  15& 186& 122&  47&  40&  23&  14 \\
   &       & 2     & 1     &         & 133&  74&  33& 219& 164&  66&  36&  47&  24 \\
   &       &       & 2     &         &  25&  15&   1&  54&  40&  13&  14&   6&   4 \\
   & 2     & 1     & 1     &         & 228& 153&  60& 356& 343& 166&  95&  80&  41 \\
   &       &       & 2     &         &  75&  45&  12& 125& 116&  34&  25&  20&  12 \\
   &       & 2     & 1     &         &  41&  25&  13&  64&  56&  22&  15&  14&  11 \\
   &       &       & 2     &         &  17&   6&   1&  19&  18&   6&   3&   3&   2 \\ \hline
\end{tabular}
\end{table}
The following example concerns a larger contingency table
including two ordinal variables with three levels.
In the analysis these variables
are treated as nominal variables using the baseline contrasts \eqref{eq.c-matrix}.
Although the nature of the variables could be handled by using other more appropriate
contrasts, as explained in  \cite{bar-col-for:2007}, the fit of the marginal
independence model is nevertheless invariant.
\begin{example}
Table~\ref{tab.usgss} summarizes observations for 13067
individuals on 6 variables obtained from as many questions taken
from the U.S. General Social Survey \citep{gss:2007} during the
years 1972-2006. The variables are reported below with the
original name in the GSS Codebook:
\begin{itemize}
\item[$C$]
\textsc{cappun}: do you favor or oppose death penalty for persons
convicted of murder? (1=favor, 2=oppose)
\item[$F$]
\textsc{confinan}: confidence in banks and  financial institutions (1=   a great deal,
2=  only some, 3=  hardly any)
\item[$G$]
\textsc{gunlaw}:  would you favor or oppose a law which would require a person
           to obtain a police permit before he or she could buy a gun? (1=favor, 2=oppose)
\item[$J$]
\textsc{satjob}: how satisfied are you with the work you do? (1 = very satisfied,
               2=  moderately satisfied, 3 = a little dissatisfied, 4=  very dissatisfied).
Categories 3 and 4 of \textsc{satjob} were merged together.
\item[$S$]
\textsc{sex}: Gender (f,m)
\item[$A$]
\textsc{abrape}: do you think it should be
           possible for a pregnant woman to obtain legal abortion if
           she became pregnant as a result of rape? (1= yes, 2 = no)
\end{itemize}
\begin{figure}[b]
\centering
\begin{tabular}{cc}
\includegraphics[height=2.5cm]{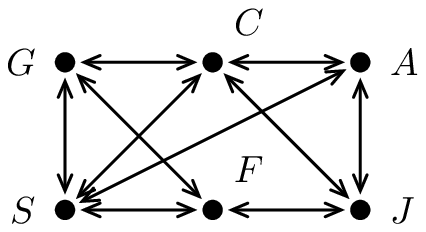} &
\includegraphics[height=2.5cm]{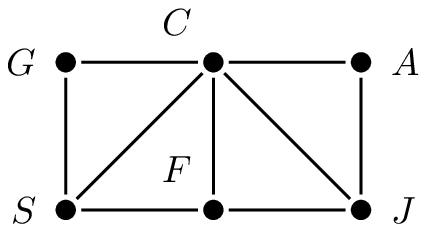} \\
(a) & (b)  \\
\end{tabular}
\caption{\it Data from the U.S. General Social Survey 1972-2006.
(a) A bi-directed graph model ($\chi^2_{17} = 17.29$).
(b) A graphical log-linear model ($\chi^2_{110} = 103.16$).}
 \label{fig.gss.usa}
\end{figure}
In data sets of this kind there are a large number of missing values and the
table used in this example collects only individuals with complete observations.
Therefore, the following exploratory analysis is intended to be only  an
illustration with a realistic example.
From a  first analysis of the data, the  following
marginal independencies are not rejected by the chi-squared
goodness of fit test statistic
$$
\begin{matrix}
F \ci CA &  G \ci JA &   J \ci GS & A \ci FG\\
\chi^2_6 = 6.7 & \chi^2_5 = 3.3 & \chi^2_6=8.1 & \chi^2_5=2.1\\
\end{matrix}
$$
and thus they suggest the independence model represented by the bi-directed graph
in Figure~\ref{fig.gss.usa}(a).
Fitting this model, under the multinomial sampling assumption,
we obtain an adequate fit with a deviance of $17.29$ on 17 degrees of freedom.
The Aitchison and Silvey's algorithm converges after 13 iterations.
The encoded independencies cannot be represented by a directed acyclic graph model
with  the same observed variables, because the graph contains at least one  subgraph
which is a chordless 4-chain. The disconnected set parameterization
defined by the ordered decomposable sequence
$$
\mcal_G = \{ CF, FA, GJ,  GA,  JS,  CFA, FGA, GJS, GJA, CFGJSA\}
$$
is variation independent.
Instead, by searching in the class of graphical log-linear
models with the backward stepwise selection procedure of \textsc{mim} \citep{edwards:2000}
we found a model with a deviance of 103.16 over 110 degrees of freedom. The model
graph is shown in Figure~\ref{fig.gss.usa}(b).
Other selection procedures show however that there are several equally well
fitting models. The chosen undirected graph is slightly simpler
(2 edge less) than the bi-directed graph. As anticipated, the number of constraints
on parameters is however much higher.
From the inspection of  the studentized multivariate logistic  estimates,
we noticed that the higher order log-linear parameters are almost all
not significant and thus we fitted a reduced model, by further
restricting to zero  all the log-linear parameters of order higher than two,
obtaining a deviance of $108.34$ on $118$
degrees of freedom. The estimates of the remaining nonzero two-factor log-linear parameters  are
shown in Table~\ref{tab.estim.gss}. These are estimated local log odds-ratios
in the selected two-way  marginal tables and they have the expected
signs.
By comparison, the fitted  non-graphical log-linear
model with the  graph of Figure~\ref{fig.gss.usa}(b), with additional
zero constraints on the log-linear parameters of order higher than two, leads
to a chi-squared goodness of fit  of $118.49$ on $119$ degrees of freedom.  Both models
thus appear adequate.
\begin{table}[t]
\centering
\caption{\it Estimates of two-factor log-linear parameters for the bi-directed graph
model of Figure~\ref{fig.gss.usa}(a) with  additional zero restrictions on higher order
terms. The asterisks indicate the parameters for which
the Wald statistic is significant.}\label{tab.estim.gss}
\small
\begin{tabular}{ccrrcccrrc} \hline
Margin & Parameter  &         Estimate   &  s.e. &&  Margin & Parameter  &         Estimate  & s.e.& \\ \hline
$CG$ & (1)       &          $  -0.38$ & $0.048 $ &*&$FG$ & (1)       &          $  -0.01$ & $0.047 $&\\
$CJ$ & (1)       &          $   0.10$ & $0.043 $ &*&     & (2)       &          $   0.16$ & $0.058 $&*\\
     & (2)       &          $   0.14$ & $0.058 $ &*&$FJ$ & (1)       &          $   0.29$ & $0.044 $&*\\
$CS$ & (1)       &          $   0.46$ & $0.040 $ &*&     & (2)       &          $   0.05$ & $0.065 $&\\
$CA$ & (1)       &          $   0.56$ & $0.049 $ &*&     & (3)       &          $   0.04$ & $ 0.056 $&\\
$GS$ & (1)       &          $  -0.77$ &$0.042 $ &*&     & (4)       &          $   0.36$ & $ 0.072 $&*\\
$JA$ & (1)       &          $  -0.21$ & $0.051 $ &*&$FS$ & (1)       &          $  -0.00$4& $0.040$&\\
     & (2)       &          $  -0.03$ & $0.075 $ &&     & (2)       &          $  -0.35$ & $0.051$&*\\
$SA$ & (1)       &          $   0.18$ & $0.047 $ &*&     &           &                            &  \\ \hline
\end{tabular}
\end{table}
\end{example}
The last example shows that sometimes the best fitted marginal
independence model may be simpler than the best fitted directed acyclic model.
\begin{example}
The set of data in Table~\ref{tab.hsss} is taken from the General Social Survey in
Germany in 1998 \citep{allbus:1998}. In a selected population aged between 18 and
65, the answers of 1228 respondents are collected about the
following 5 binary variables
$U$, unconcerned about environment (yes, no); $P$, no own political impact expected (yes, no),
$E$; parents education, both at lower level (at most 10 years) (yes, no);
$A$, age under 40 years(yes, no); $S$, gender (female, male).
\begin{table}[b]
\caption{\it Data from the German  General Social Survey in 1998.} \label{tab.hsss}
\small
\centering
\begin{tabular}{lllrrrrrrrr} \hline
 &  & $U$ &  yes &    &    &    &  no &   &    &  \\
 &  & $S$ &  f &    &  m &    &  f &   &   m&    \\ $A$& $E$& $P$ &  yes &  no &  yes &  no &  yes &  no&   yes&   no  \\     \hline
no& yes&   &  6 &  8 &  7 & 27 & 66 &186&  24& 230    \\
 & no&   &  4 &  0 &  1 &  9 &  8 & 64&   4&  60    \\
yes& yes&   &  2 &  2 & 11 &  6 & 28 &159&  16& 130    \\
 & no&   &  0 &  1 &  0 &  2 &  4 & 75&   8&  80    \\ \hline
\end{tabular}
\end{table}
A possible  ordering of the variables has been suggested
by \cite{wermuth:2003}, who analyzed a superset of this data set
and discussed a directed acyclic graph model. Using a similar ordering,
limited to  the variables here studied, we consider the  variables  $\{A,S\}$ as
purely explanatory, $E$ and  $P$ as intermediate and $U$ as final response.
Our final well fitting directed acyclic graph model, shown in
Figure~\ref{fig.GSS}(a), has a deviance $3.70$ over 3 degrees
of freedom. The subgraph for all the variables except gender $S$
is complete. Specifically, the graph has an edge $E \rightarrow U$,
indicating a direct effect of education on the final response. The
model without the arrow $E \rightarrow U$ has a worse goodness of fit
$\chi_{15}^2 = 36.0$ and
further it can be verified that the two-factor log-linear parameters $EP$ and $EA$
are large and significant.
 Model selection in the
class of the graphical log-linear  models does not lead to any sensible
reduction whilst search in the class of bi-directed graph models shows that
a special structure of marginal independencies holds.
The final selected bi-directed graph, represented in Figure~\ref{fig.GSS}(b),
represents  the marginal independencies $S \ci A,E$ and $E \ci S,U$.
The bi-directed graph contains the chordless
4-chain $EAUS$ and thus it is not Markov equivalent to any directed
acyclic graph
in the five variables. This
suggests that the directed acyclic graph model conceals some
distortions due to the presence of latent variables.
\begin{figure}[t]
\centering
\begin{tabular}{ccc}
\includegraphics[height = 2.4cm]{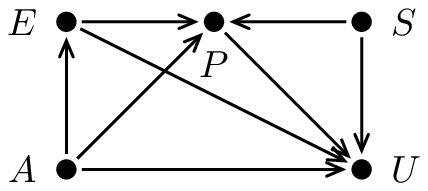} &
\includegraphics[height = 2.4cm]{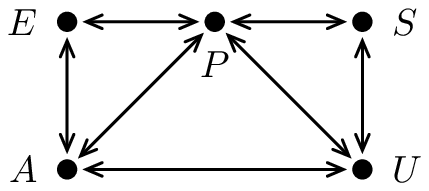} \\
(a) & (b)  \\
\end{tabular}
 \caption{\it Two graphical models fitted to data
from the General Social Survey in Germany, 1998. (a) A directed
acyclic graph  model: $\chi^2_3 = 3.70$. (b) A bi-directed graph model:
 $\chi_{5}^2 = 5.91$.}\label{fig.GSS}
\end{figure}
Also in this case, the disconnected set parameterization
defined by the sequence $\mcal_G = (GE, GF, AE, GFE, GEA, ABEFG)$
leads to a variation
independent parameterization because it can be verified that
the sequence $\mcal_G$ is order decomposable.
\end{example}
\section{Discussion}\label{sec.dis}
The discrete models based on marginal log-linear models by
\citet{ber-rud:2002} form a large class that includes several
discrete graphical models. The undirected graph models and the
chain graph models under the classical (Lauritzen, Wermuth,
Frydenberg) interpretation can be parameterized as marginal
log-linear models. For an introduction see
\citet{rud-ber-nem:2006}. This paper shows that the discrete
bi-directed graph models under the global Markov property are
included in the same class by specifying the constraints
appropriately. In general, three main criteria were considered in
choosing a marginal log-linear parameterization.
\begin{itemize}
\item[(a)] Upward compatibility:  if the parameters have a meaning
that is invariant across different marginal distributions, then
the interpretations remain the same when a sub-model is chosen. We
saw that the multivariate logistic parameterization has this
property. \item[(b)] Modelling considerations: the
parameterization should contain all the parameters that are of
interest for the problem at hand.  For example, in a regression
context where some variables are prior to others, effect
parameters conditional on the explanatory variables are most
meaningful. In the seemingly unrelated regression problem of
Example~\ref{sur}, the chosen parameters have the interpretation
of logistic regression coefficients. \item[(c)] Variation
independence: if the parameter space is the whole Euclidean space,
this has certain advantages. First, the interpretations are
simpler, because in a certain sense different parameters measure
different things. Second, in a Bayesian context, prior
specification is easier. Finally, the problem of out-of-bound
estimates when transforming the parameters to probabilities is
avoided. In the examples, we always found a variation independent
parameterization, but a characterization of the class of
bi-directed graphs admitting a variation independent complete and
hierarchical marginal log-linear parameterization is an open
problem.
\end{itemize}
The three criteria are in some cases conflicting: typically variation independence
is obtained at the expense of upward compatibility.

The multivariate logistic parameterization has a purpose similar to that of the
 M\"obius parameterization  recently proposed by
\citet{drt-ric:2007} for binary marginal independence models,
which is based on a minimal set of  marginal probabilities
identifying the joint distribution.  These authors discuss  the type
of constraints on the M\"obius parameters needed
to specify a marginal independence, showing that they take a simple
multiplicative form. The same constraints are defined by zero restrictions
on marginal log-linear  parameters in our approach. Even if the
parametric space can be awkward, this problem
is handled by a  fitting algorithm that
operates in the space of the expected frequencies, while
the parameters are used only to define the independence constraints.
Moreover, the definition of the models through the complete specification
of the marginal log-linear parameters  gives some advantage
when there is a mixture of nominal and ordinal variables
because it allows  to define appropriate  parameters for
both types of variables using the theory of  generalized marginal
interactions by \citep{bar-col-for:2007}.  This opens the
way to defining subclasses of discrete graphical  models
specifying equality and inequality constraints.

The proposed algorithm for maximum likelihood fitting of  the
bi-directed graph model is a very general algorithm of constrained
optimization based on Lagrange multipliers. It is essentially
based on \citet{ait-sil:1958} as later developed by
\citet{ber:1997}. Similar algorithms have been proposed, for
instance, by \citet{mol-les:1994}, \citet{glo-mcc:1995},
\citet{lang:1996} and further generalized by \citet{col-for:2001}.
Its main advantage is its generality (it can be applied to all
models defined by constraints on the marginal log-linear
parameters). As previously stated, the algorithm does not require
further iterative procedures for computing, at each step, the
inverse transformation from the marginal log-linear parameters to
the cell probabilities. Thus, the risk of not compatible estimates
that could arise for the lack of variation independence is
avoided. The disadvantage is that, as for many gradient-based
algorithms of this type, convergence is not guaranteed and that it
requires the computation of a large expected information matrix.
However, empirically, convergence  is achieved in a relative few
number of iterations by including a step adjustment. An
alternative algorithm with convergence guarantees is the Iterated
Conditional Fitting algorithm, proposed  by \citet{drt-ric:2007}
for binary bi-directed graph models in the M\"obius
parameterization. A comparison between the  two algorithms in
terms of performance, speed and memory requirements needs further
investigation.
\section*{Acknowledgement} We thank Nanny Wermuth  for helpful discussions. The work of the first two authors
 was partially supported by MIUR, Rome, under the project PRIN 2005132307.

\bibliographystyle{sjs}
\small
\bibliography{bgm}
\end{document}